\documentclass[10pt,journal]{IEEEtran}

\usepackage[T1]{fontenc}
\usepackage[utf8]{inputenc}
\usepackage{lmodern}
\usepackage{microtype}

\usepackage{graphicx}
\usepackage{booktabs,tabularx,array}
\usepackage{xcolor}
\usepackage{stfloats}
\usepackage{enumitem}
\usepackage{url}

\usepackage[
    backend=biber,
    style=ieee,
    sorting=none,
    maxbibnames=4
]{biblatex}

\usepackage[hidelinks]{hyperref}

\newif\ifRAMshowabstract
\RAMshowabstractfalse

\graphicspath{{figures/}}

\definecolor{navy}{RGB}{30,56,81}
\definecolor{teal}{RGB}{20,121,125}
\definecolor{pale}{RGB}{235,242,247}
\definecolor{mint}{RGB}{232,245,241}
\definecolor{warm}{RGB}{252,243,225}
\definecolor{edge}{RGB}{25,99,130}

\newcommand{\lesson}[1]{%
  \par\smallskip\noindent
  \textbf{Design choice.}~#1
}

\newcommand{\cost}[1]{%
  \par\smallskip\noindent
  \textit{Tradeoff.}~#1
}

\newcommand{\pullquote}[1]{%
  \begin{center}
  \setlength{\fboxsep}{7pt}
  \fcolorbox{edge}{mint}{%
    \parbox{.86\linewidth}{\centering\bfseries #1}
  }
  \end{center}
}

\title{When ``No Worries'' Went Wrong:\\
HRI Lessons for Foundation-Model Robots}

\author{Eliahu~Khalastchi%
\thanks{Eliahu Khalastchi is with The College of Management Academic Studies.}%
}

\renewenvironment{IEEEkeywords}
{\par\smallskip\noindent\textbf{Keywords:}~}
{\par\medskip}

\begin{document}
\maketitle

\begin{center}
\footnotesize\itshape
This work has been submitted to the IEEE for possible
publication. Copyright may be transferred without
notice, after which this version may no longer be accessible.
\end{center}

\begin{abstract}
When a person declined our robot's invitation to play, Alice turned her shoulder away and said ``No worries, it's all good.'' Only after a brief pause did she drive off. The words and gestures were plausible but their order changed the message. In another episode, Alice informed a host that his visitor had arrived---while the visitor stood right beside him. Both came from a three-year ROS~2 HRI consortium that integrated large language and vision models as they emerged. Today's foundation models give robots broader capabilities to perceive their surroundings, reason about situations and generate behavior. Yet capability alone does not ensure that a changed situation shapes the action that reaches a person. Evidence may never be acquired, or a purpose or constraint may fail to survive planning and execution. A system may even report \texttt{Success} when an action no longer serves its intended purpose. Drawing on the consortium's engineering experience, we develop five design lessons about attention, knowledge, commitments, embodied timing and action authority. We then introduce TRACE, a lightweight review that changes one interaction premise, follows it to a decision or enforcement boundary, and produces an actionable architecture finding. A bounded analysis of a published LLM-generated behavior-tree system illustrates how an available clarification mechanism can go unused. The engineering lesson is to preserve accountability boundaries, not the computational pipelines that once exposed them.
\end{abstract}

\begin{IEEEkeywords}
human--robot interaction, social robots, robotics architecture, foundation models, multimodal integration, design patterns, engineering practice
\end{IEEEkeywords}

\section{When a Robot Accidentally Sounded Passive-Aggressive}
\begin{figure}[!htb]
\centering
\includegraphics[width=\linewidth,keepaspectratio]{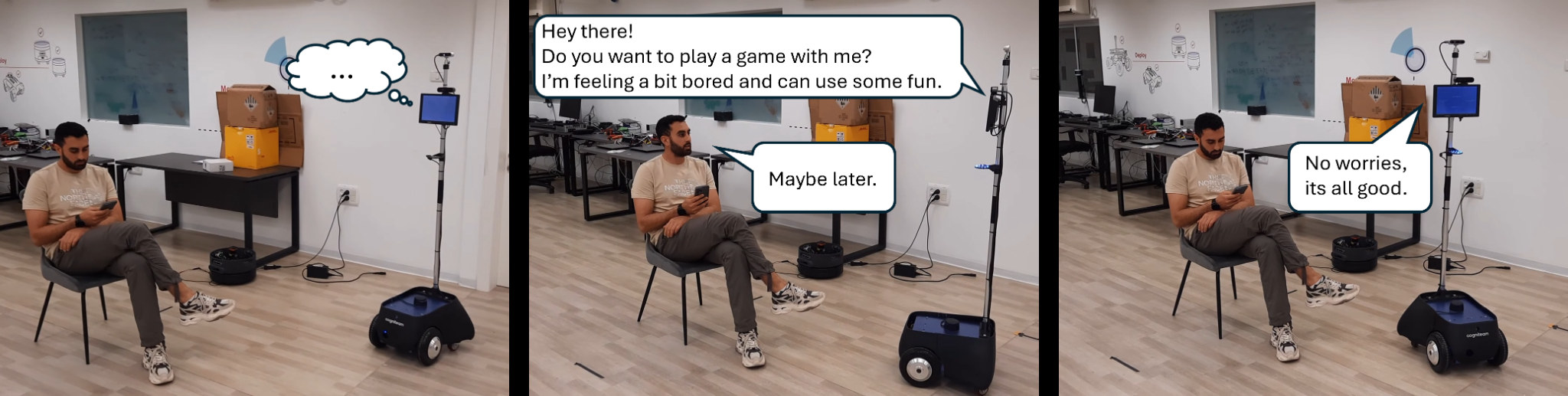}
\caption{A demonstration gone hilariously wrong: Alice turns away before saying ``No worries,'' so an asynchronous reassurance reads as passive-aggressive.}
\label{fig:no_worries_comic}
\end{figure}

A person was looking at his phone when Alice, our office robot, cheerfully invited him to play a game. ``Maybe later,'' he replied. Alice pivoted away, presenting him with her ``shoulder.'' Then she stopped and, with impeccable verbal composure, declared \emph{``No worries, it's all good.''} She waited another beat before driving away. The pause and turned posture made the reassurance sound remarkably like wounded pride. We had not programmed her to sulk (Fig.~\ref{fig:no_worries_comic}).

The explanation was engineering. Originally, JSON-configured behaviors sequenced the reply before the disengagement gesture. An LLM made responses more natural and context-sensitive but added generation latency, particularly when run locally. We allowed speech preparation and movement to proceed asynchronously to avoid an awkward wait. The orientation change then completed \emph{before} the words were spoken, even though Alice did not drive away until afterward. The separate components behaved as programmed; their ordering made the overall interaction convey something else.

An upgraded model can improve a component while changing the social effect of the complete robot. Current foundation models may generate subtler responses and richer movement, yet the executor, scheduler or tool boundary can still determine which effect reaches a person first---or whether the relevant observation is gathered at all. \emph{Foundation models expand and relocate robot capabilities; appropriate interaction still depends on acquiring the right evidence and making purpose and constraints actionable at the point of effect.} We investigate both breaks: a fact that never becomes usable evidence, and information that cannot change execution when it matters.

\section{From an Integration Challenge to a Changing AI Landscape}
The consortium began work in May 2022, after nearly two years of formation. Specialized face, gesture and skeleton recognition, speech understanding, navigation and affect estimation were already developing rapidly. The difficult part was connecting their outputs into a robot that could perceive a situation, retain human and scene context, choose a response and execute it coherently. Multiple partners needed to replace or improve their own experts without requiring a redesign of everyone else's software. To address that integration challenge, we built a reusable ROS~2 architecture, the \emph{HRI Toolkit}.

ROS4HRI had introduced interoperable human-centered ROS interfaces at IROS 2021 and through subsequent conventions \cite{mohamed2021ros4hri,ros4hri2022rep}. The consortium addressed a complementary problem: composing capabilities into an interaction loop that extended from perception and context to social planning and embodied response.

\begin{figure*}[!t]
\centering
\includegraphics[width=\textwidth,keepaspectratio]{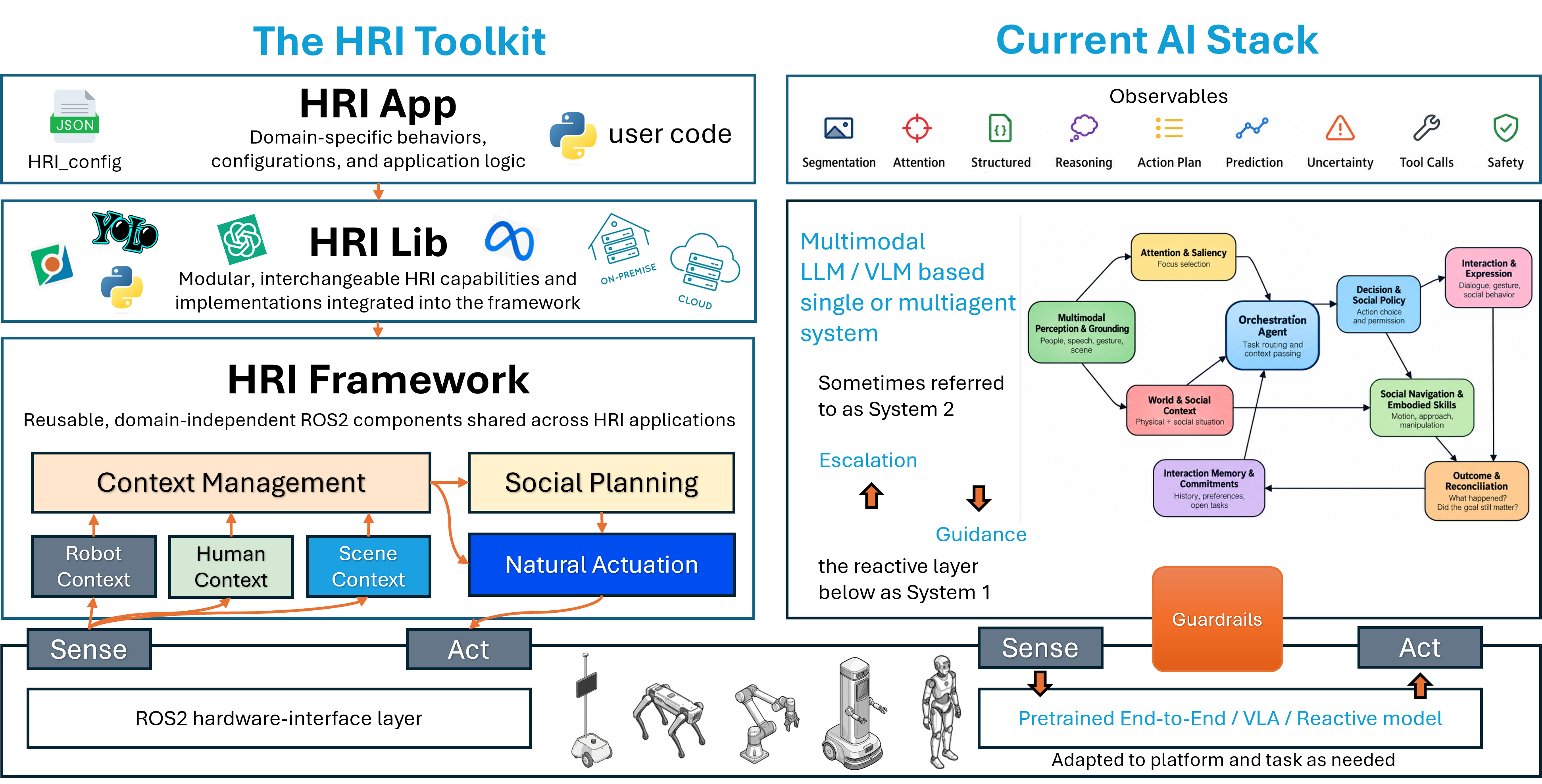}
\caption{Historical HRI Toolkit (left) beside one possible current Physical AI design space (right). The former makes modular processing responsibilities explicit; the latter depicts possible deliberative and reactive paths, guardrails and observable outputs. Robot silhouettes are illustrative, not a count of tested platforms.}
\label{fig:architecture_comparison}
\end{figure*}

The left side of Figure~\ref{fig:architecture_comparison} presents the Toolkit's three levels. The \emph{HRI Framework} supplied reusable, domain-independent ROS~2 responsibilities for robot, human and scene context, context management, social planning and natural actuation. The \emph{HRI Library} packaged interchangeable experts and implementations. Teams could switch, for example, between object-detection or language models, and between edge, on-premises and cloud execution; in one demonstration, a spoken instruction to the robot switched the conversation model and language at runtime without rebuilding the surrounding application. The \emph{HRI App} defined domain-specific behavior, configuration, policies and user code. Commands marked as reactive, such as ``stop,'' could bypass social planning.

The Toolkit was exercised on several platforms, including social robots, quadrupeds and manipulators. We deliberately used non-humanoid forms: the aim was behavior perceived as natural without relying on a human shape. The particular robot shape mattered less than whether the same interaction responsibilities could be reused. Named interfaces made errors localizable: developers could inspect a context field, an event, a handler or an expert's input and output, and when and why it was selected. The retrospective design diagrams below explain how we came to express those responsibilities.

\begin{figure}[!htb]
\centering
\includegraphics[width=.78\linewidth,keepaspectratio]{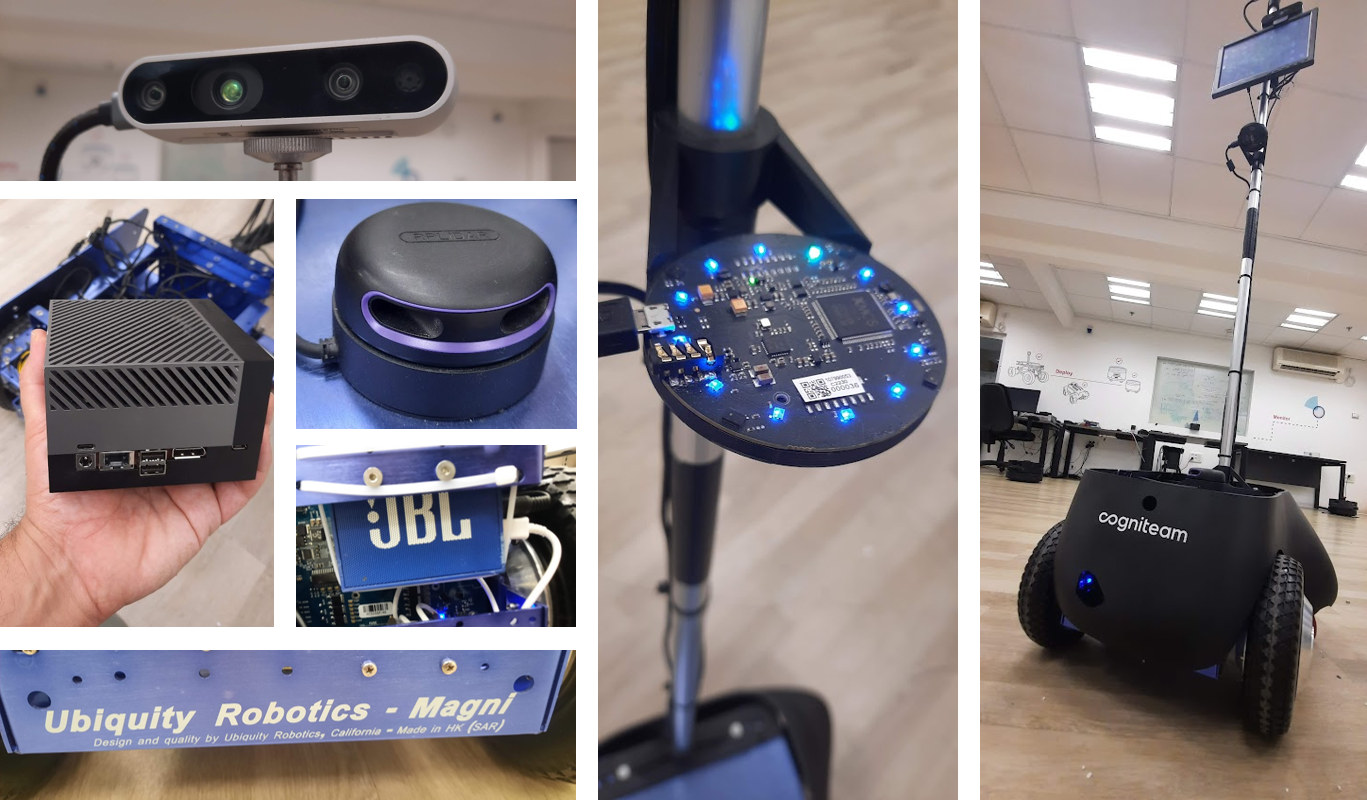}
\caption{Alice, the office robot built by cogniteam for the consortium. The Magni mobile base carried an Intel RealSense depth camera, 2D LiDAR, a Jetson onboard computer, and a microphone array supporting sound-direction estimation. Collage from the project hardware and demonstration archive.}
\label{fig:robot}
\end{figure}

\subsection{One Person, Many Experts}
\begin{figure}[!htb]
\centering
\includegraphics[width=.92\linewidth,keepaspectratio]{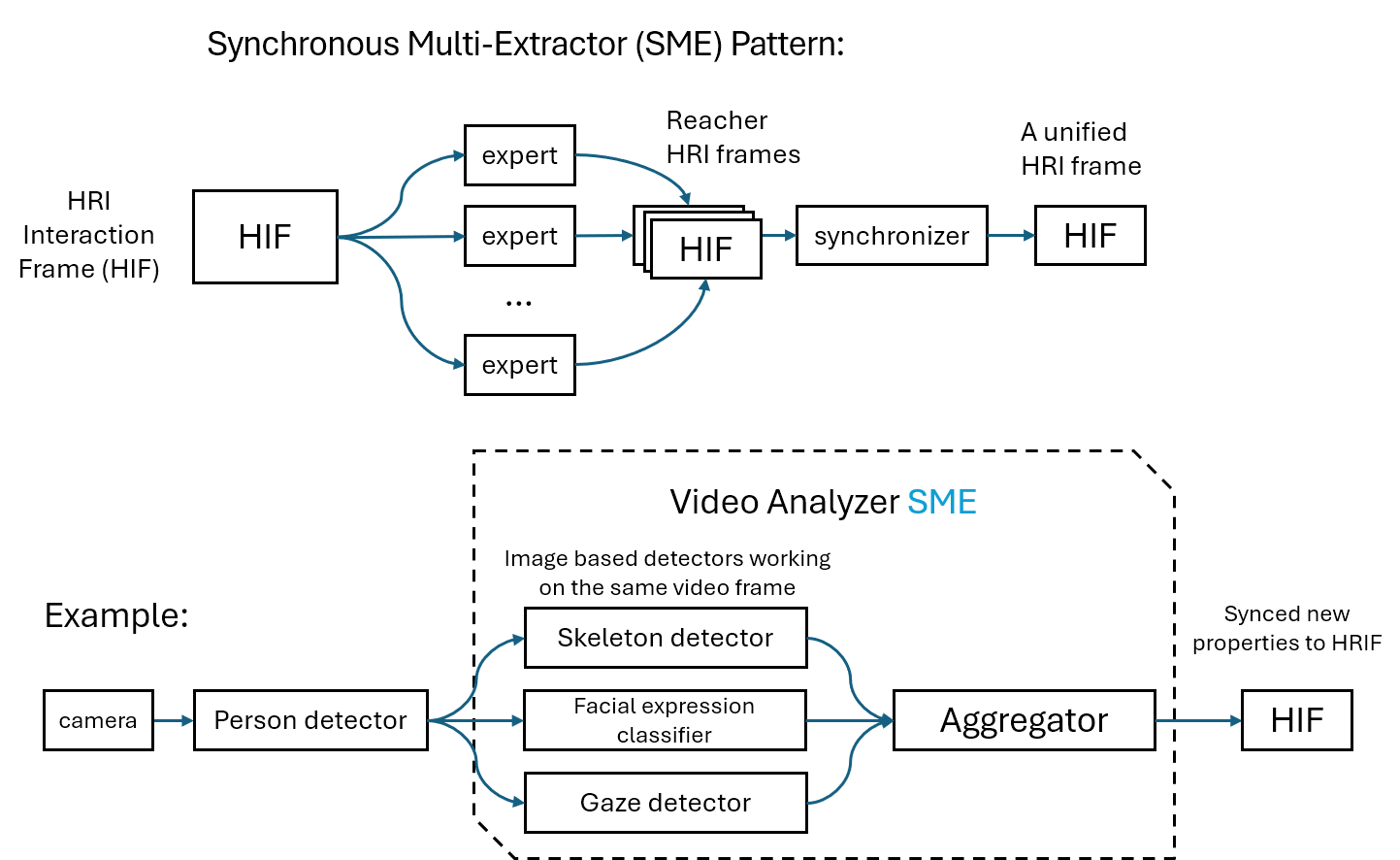}
\caption{Retrospective parallel expert enrichment: visual experts operate on the same observation, then a synchronizer produces an enriched HRI Interaction Frame (HIF). The lower panel illustrates video analysis.}
\label{fig:sme}
\end{figure}

In robots such as Alice (Fig.~\ref{fig:robot}), a single observation could feed several experts. Figure~\ref{fig:sme} shows experts estimating skeleton, facial expression and gaze for the same person, then synchronizing their contributions. In the retrospective S.O.C.I.A.L. modeling catalog \cite{khalastchi2026site}, each enriched HRI frame can act as an \emph{interaction passport}, retaining properties, sources, timestamps and confidence. The patterns were abstracted from the Toolkit's code and formalized afterward; deployed nodes did not literally exchange HML frames.

\begin{figure}[!htb]
\centering
\includegraphics[width=.96\linewidth,keepaspectratio]{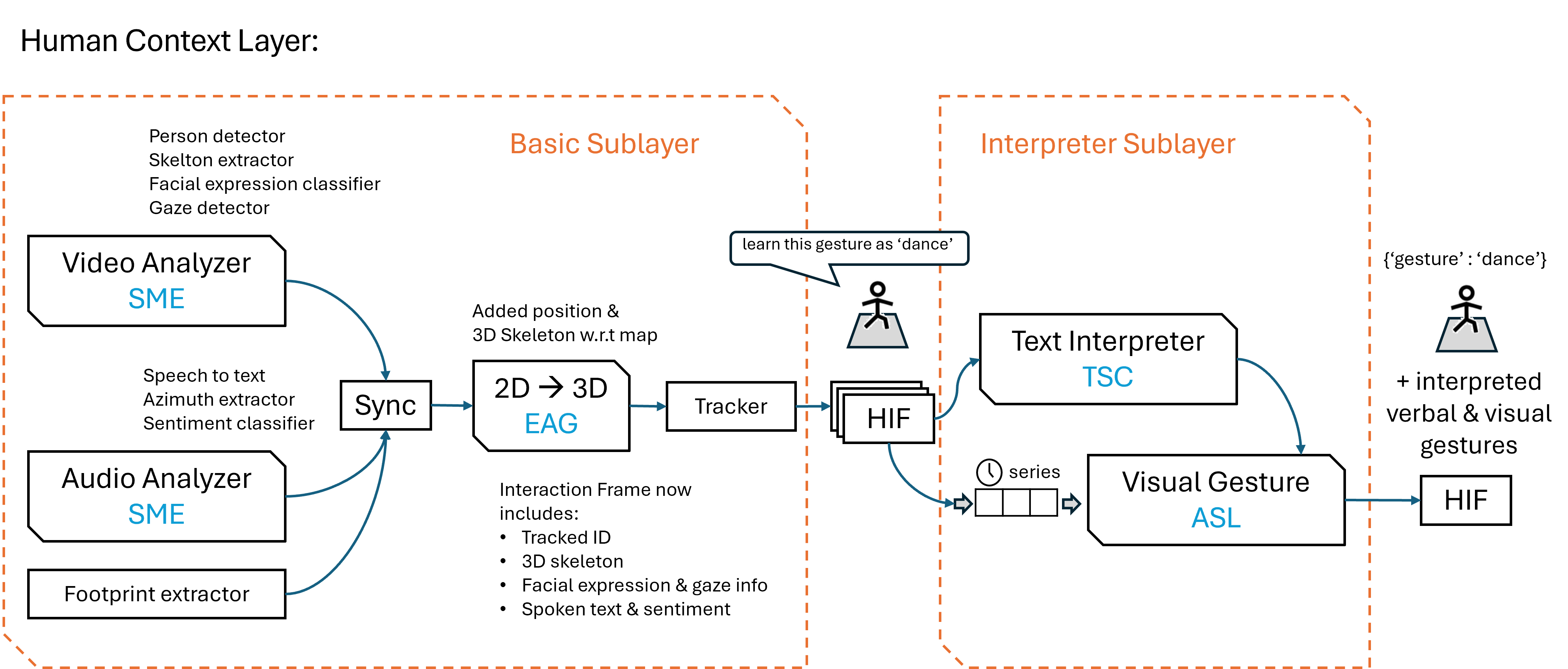}
\caption{Retrospective Human Context composition of four generic patterns: SME synchronizes analyses; EAG allocates processing effort; TSC provides tiered language interpretation; ASL learns user-specific temporal signatures, illustrated here as gestures.}
\label{fig:full_hc}
\end{figure}

Figure~\ref{fig:full_hc} shows four such patterns in Human Context. Each is \emph{generic}, applied and reused in different contexts. The SME pattern (Fig.~\ref{fig:sme}) is applied to video and audio analysis. An \emph{Elastic Attention Governor} (EAG) prioritizes and allocates processing by relevance, uncertainty and resources; here, which people receive 2D-to-3D enrichment first, and by which expert. A \emph{Tiered Semantic Cache} (TSC) escalates from cheap to costly interpretation and caches results; here, from NLP templates to an LLM (Section~\ref{sec:shift}). An \emph{Adaptive Signature Learner} (ASL) learns recurring temporal patterns; here, user-specific gestures. The catalog \cite{khalastchi2026site} shows how these and further patterns combine into the full architecture.

The tradeoff is that explicit experts exposed \emph{which evidence was selected, where it came from, when it arrived, and how it contributed}---at the cost of interfaces, synchronization and code. These responsibilities remain even as their implementation increasingly dissolves into learned weights.

Earlier HRI design patterns examined social interactions \cite{kahn2008designpatterns,sauppe2014designpatterns}; this later catalog focused on software responsibility boundaries that may now be hidden within models.

\section{The Ground Shifted While We Were Building}\label{sec:shift}
The consortium architecture absorbed new models as they became useful, starting with emerging language models in late 2022. Their roles expanded through 2025; three uses proved instructive.

\textbf{Free language, approved actions.} Initially, speech-to-text fed a conventional NLP interpreter recognizing patterns with parameters, such as \texttt{where is <who>} and \texttt{bring the <object> to <person>}. The application's JSON handlers defined the corresponding permitted behaviors. We added an LLM that mapped varied expressions---for example, ``Do you know Bob's whereabouts?''---to an existing template and its argument. The rest of the pipeline remained unchanged. Cached mappings reduced future inference time, although an incorrect mapping could also become a persistent error. This design intentionally limited execution to the application's approved repertoire; an unfamiliar request could be declined instead of converted directly into arbitrary API calls.

\textbf{Reasoning over explicit knowledge.} Our HRI\_DB held people, objects, locations, relationships and preferences as nested dictionaries readily serialized to JSON. Graph-style queries evolved toward LLM-assisted reasoning, including LLM+Prolog and later direct LLM reasoning. The model could connect ``bring me the bottle in my favorite color'' to stored preferences and bottle attributes, or reason over a person's relative location. Developers could still inspect and, when needed, correct the records used as premises.

\textbf{Flexible expression within bounded behavior.} LLMs rewrote accepted replies for conversational appropriateness. The ``No worries'' episode revealed the new latency and sequencing costs.

Over time, models and services were replaced behind the same interfaces without changing the application, including vision-language models as they matured.

A boundary around authority remained deliberate. In a non-robot demonstration, asking ChatGPT for a behavior tree to ``ram Bob'' yielded a plausible approach-and-drive sequence. This highlighted the risk of granting generative suggestions unchecked physical authority. We used models as interpreters, reasoners and stylists while approved handlers committed behavior; later experiments also permitted model-proposed handler code subject to developer approval. Related work explored grounded language-to-skill selection and language-generated control programs \cite{ichter2023saycan,liang2023cap}.

Those handlers offered predictable scope at the price of manual construction and limited generality. Modern learned controllers can remove much of that work. The lasting questions concern the \textbf{responsibilities} that those handlers had forced us to name.

\section{``The Consortium Manager Is Here!''}
On the evening before a consortium meeting, we loaded photographs of expected visitors so that Alice could recognize and greet them. The next morning, the host asked Alice in the kitchen to let him know when the consortium manager arrived. The host then left the kitchen. The manager arrived, and the two were already talking when, about ninety seconds later, Alice approached, turned toward the host and solemnly informed him that the consortium manager had arrived. The manager was standing right beside him. We laughed; before we had fully taken in the absurdity, Alice turned to the manager and asked about his grandchildren. The information was correct, but its timing made the message wonderfully redundant.

The system detected the event and completed a pending action. It had no implemented representation of \emph{why} notifying the host might be useful, no rule translating an observed encounter into evidence that the likely purpose had been satisfied, and no last-moment decision about whether the notification still mattered. A modern model might identify two people conversing, but that interpretation helps only if it is actually acquired, reaches the relevant decision point and is permitted to alter the action.

If the message is meant to make the host aware of the manager's arrival, the robot should reconsider it when that outcome may already have been achieved. A request for a formal announcement, however, could justify delivering it regardless. Under the first policy, the historical implementation had a purpose-reassessment gap; under the second, there is no violation. An inferred purpose cannot silently cancel an explicit instruction. Delivery, deferral or clarification remain policy-dependent alternatives. This is a social analogue of the time-of-check/time-of-use (TOCTOU) problem familiar from software engineering \cite{bishop1996race}, compounded by lost purpose.

\pullquote{Decide late enough, and remember why.}

\section{The Model-Centric HRI Stack: Richer Capabilities, New Handoffs}
Would a foundation-model robot have avoided Alice's redundant notification? Possibly---but only if its interpretation could influence the pending action. Here, \emph{Physical AI} means AI systems that interpret and act in the physical world through embodied sensing and control. Generalist robot policies such as RT-2 and OpenVLA can map visual and language inputs into action \cite{zitkovich2023rt2,kim2025openvla}. Language-driven multimodal HRI, model-generated programs and behavior trees, and combinations of embodied reasoning with VLA execution offer other useful designs \cite{wang2024lami,liang2023cap,merino2025bt,gemini2025robotics}. The right panel of Fig.~\ref{fig:architecture_comparison} sketches the modern design space alongside our historical Toolkit.

A fast \emph{reactive or learned policy} (``System~1'') may act directly on observations, while deliberation (``System~2'') may run in one model, several agents or ordinary software; hybrids escalate hard cases and send guidance back. A fast VLA is not automatically unobservable, nor is a multi-agent system automatically explainable.

\textbf{Observability is no longer synonymous with an explicit expert pipeline.} The right panel of Fig.~\ref{fig:architecture_comparison} depicts example interfaces such as segmentation, attention, structured output, action plans, predictions, tool calls and safety checks; others expose retrieval provenance or commitment states. Their evidential strengths differ: attention maps and model-reported reasoning are not guaranteed faithful causal explanations \cite{jacovi2020faithful,turpin2023unfaithful}, whereas actual tool calls, timestamps, enforcement decisions and state changes offer stronger operational traces.

\emph{The key distinction is between what a model can do and what the system lets it do at the moment that matters.} A VLM may be able to identify a conversation but never be consulted when a notification fires. A language model might reason correctly over a preference but receive stale evidence. A socially apt utterance can still arrive after its intended gesture. Two routes to failure therefore matter: \emph{the changed premise is never acquired as usable evidence}, or \emph{it is available but fails to influence the authorized effect}. Even a unified end-to-end controller needs sufficiently current evidence; in compositional systems, the component that interprets the scene need not be the one that commits its consequence. TRACE follows the premise to the boundary where it can change the effect.

Could learned models acquire these behaviors from data? Many can learn social conventions and expressive timing without handcrafted rules. Yet human feedback can be ambiguous: laughter may accompany an awkward mistake rather than approval. Nor is a local policy or current situation necessarily present in the training set. Even a model that knows the norm needs the relevant evidence when acting. The engineering aim is to expose and check consequential premises where needed, not to replace learned behavior with a rigid pipeline.

\section{Five Responsibilities We Can No Longer Take for Granted}
The shift brings five previously explicit responsibilities into focus. They are distinct questions, not pipeline stages; one incident may involve several.
\subsection{L1. Treat Attention as a Budget}
Our gesture recognizer initially had to process irrelevant motion in the background, generating spurious candidate events and wasting computation. We used inexpensive cues, including proximity and body orientation, to prioritize interaction candidates; in a live stage demonstration, we restricted perception to the stage rather than treating the entire auditorium as equally relevant. Attention also has a conversational dimension: Alice could respond with ``Hi! How can I help?'' on hearing her name in a discussion \emph{about} her, although nobody had addressed her.

A modern multimodal model may replace several detectors yet still faces finite context, latency and compute budgets. The engineering question is which people, regions, utterances and events become candidates for interaction. Whether selection is handled by a separate component or a learned policy, developers need an appropriate observable to diagnose why a candidate was selected or ignored.
\lesson{Make selection of interaction targets an inspectable responsibility: expose the candidates considered, selected or ignored, without mandating a separate attention agent.}
\cost{Aggressive selection saves computation and reduces false triggers but can hide a peripheral event that becomes urgent.}

\subsection{L2. Treat Stored Knowledge as Evidence, Not Fact}
When a question could not be answered from short-term memory or HRI\_DB, the toolkit could consult an LLM's internal knowledge rather than a live web search. Asked who the U.S. president was, the robot returned an answer qualified by the model's knowledge date. A consistency-checking component detected a conflicting user statement and asked for confirmation before updating the persistent record; later we also explored LLM-based consistency checking. The next response came from memory rather than another model call. Confirmation corrected a \emph{record}; it did not independently establish the truth of an external claim.

The issue extends beyond model knowledge cutoffs: a statement's \emph{epistemic status} depends on its source and domain. A person's preference can be represented as self-reported; a claim about a current officeholder calls for current verification; a physical door state calls for recent sensor evidence. Hypothetically, ``Charlie said the door was locked ten minutes ago'' does not license ``the door is locked.'' Permission to \emph{write} persistent memory is a separate action-authority decision (Lesson~5).

We also kept person-specific gesture preferences in HRI\_DB: our Social Reinforcement Learner adjusted the probabilities of alternative expressions from verbal or nonverbal feedback observed within a time window. The preferences persisted beyond the conversation rather than depending on temporary model context; an inferred preference from a smile or praise is nevertheless weaker evidence than a preference stated explicitly.
 Modern retrieval and tool use can provide fresher evidence. Verify in proportion to how a claim will be used: if checking would derail a low-stakes exchange, Alice could instead qualify a tentative answer (``I believe...'') and offer to check further---a more natural response than demanding confirmation for every uncertain fact.
\lesson{Attach source, scope, timestamp, confidence and verification status to claims; distinguish correcting a stored record from independently establishing an external fact.}
\cost{Provenance checks and source selection add latency and operational complexity, but silent promotion of testimony to fact can have physical consequences.}

\subsection{L3. Preserve Task Purpose as Well as Task Triggers}
The toolkit could hold a task while required information was missing, and reactivate it when the prerequisite appeared. For example, a request to \emph{bring the bottle to Bob} could wait until Bob was located, either through perception or a reply to a follow-up question, and then resume. However, pending tasks were not governed by a complete lifecycle: time-based decay was sometimes used to forget them, and recognizing that a task had become unnecessary was not generally implemented. The visitor story shows why readiness alone is insufficient.

For each consequential request, an architecture should distinguish the \emph{requested operation}, its \emph{explicit or inferred desired outcome}, and its \emph{commitment status}. When evidence suggests the outcome was achieved, withdrawn or made impossible, policy decides whether to execute, replan, wait or clarify. In an LLM-based system, an open commitment may also vanish from the working context as conversations grow, summaries replace history or plans are regenerated. Its purpose should not disappear. Recorded outcomes may also supply learning signals for when to act or clarify, provided that task completion is distinguished from mere execution success. Readiness answers \emph{can this execute?}; commitment state addresses \emph{should this still be pending under the agreed policy?}
\lesson{Retain a lightweight task-state record outside ephemeral plans, and name the events that justify completion, cancellation, expiry or clarification.}
\cost{Explicit lifecycle management consumes state and policy effort; simply timing out all tasks risks forgetting an obligation that still matters.}

\subsection{L4. Check Social Effects at Execution, Including Their Order}
A request to bring coffee once led our robot along an office corridor immediately after it had been mopped. The cleaner was understandably unhappy with the tracks it left. The robot did \emph{not} have a detector for freshly washed flooring: an \emph{acquisition gap}---the relevant evidence was never gathered, rather than lost in a handoff. We considered, but did not implement, using a VLM to assess whether a proposed route was socially appropriate \emph{now}---for instance, through a freshly cleaned corridor or a group taking photographs---rather than building a detector for every situation. Such judgments must also weigh urgency: disrupting a cleaner for a routine coffee delivery differs from delivering emergency equipment. Social navigation and VLM-assisted scoring already explore this space \cite{mavrogiannis2023navigation,song2025vlmsocialnav}.

The opening ``No worries'' incident adds a different execution concern. Even with the right action and words, timing carries meaning: modalities may run asynchronously, but some expressive acts need explicit ordering. A modern VLA may generate richer gestures, yet an external executor can still reorder them. Expression requires an execution contract---for example, begin the reassurance before the disengagement turn, or explicitly coordinate both as a single expressive act---not merely a correct utterance and a correct motion command.
\lesson{At the consequential boundary, review social context, required evidence and cross-modal ordering before committing an effect.}
\cost{Independent operations can overlap to avoid delay; only socially meaningful dependencies need ordering or selective rechecking. Excessive blocking can itself make interaction awkward.}

\subsection{L5. Separate Proposing from Committing}
Our Toolkit placed an application-configurable \emph{Decision Helper} between incoming HRI events and their Event Handlers. Rather than merely recommending a response, this injected Python component could let an event proceed, adjust its parameters, redirect it to another handler, or refuse it, returning a decision and rationale. One example was a politeness checker. A polite request could produce faster speech and livelier movement, conveying enthusiasm; an ordinary request proceeded normally. An impolite request could slow speech or trigger a refusal gesture---backing away while shaking the robot's body from side to side---along with a request to speak more politely. Agreeing to everything can feel mechanical; a context-appropriate refusal may feel more natural. Yet that policy would be misplaced in an emergency domain, where an urgency-oriented helper might be injected instead. The value was \emph{domain-specific behavioral control at runtime}, without replacing perception or action implementations.

A second, harder boundary was the approved-handler repertoire (Section~\ref{sec:shift}): an early, intentionally restrictive guardrail that blocked out-of-repertoire actions but was brittle and laborious to extend.

Modern counterparts include system prompts, injected skills, retrieved policies, tool routing, and executable guardrails. Many adapt behavior at runtime without retraining; others require fine-tuning, with regression-testing costs and forgetting risks. The goal is to change domain policies without unintended effects elsewhere. Prompts may suffice for soft preferences, while consequential actions may warrant independently testable, enforceable boundaries. Existing structured-output validators can check purpose, addressee, evidence provenance and age, permissions, ordering, and intended outcomes---though model-generated fields alone are not evidence. We recommend runtime-injectable policies that \emph{genuinely influence execution} and, when properly implemented, preserve modularity of responsibility even as computation becomes learned. \textbf{A unified model does not require unified responsibility.}
\lesson{Name the proposer, decision owner and effect owner; retain a separately reviewable, domain-appropriate route to change or block consequential effects.}
\cost{Explicit boundaries impose integration and testing effort; indiscriminate gating can suppress useful autonomy.}

\section{TRACE: Following a Changed Premise to Its Effect}
The incidents expose two failures: a relevant premise never becomes evidence, or it cannot influence the resulting effect. An \emph{accountability boundary} is an identifiable decision or enforcement point where relevant evidence, purpose, permission and execution constraints are considered for a consequential action, and where decision and outcome remain traceable. Different effects may warrant different checks; the last effective checkpoint is particularly important when action is delayed.

TRACE is a lightweight review used before deployment or after an incident. Trigger, Rationale, Assumptions, Constraints and Effects organize its record. The method changes a plausible premise and follows it through the system, adapting established scenario and system-theoretic control-analysis ideas \cite{leveson2018stpa}. Its useful output is a \emph{located architecture finding}: ``When [premise] changes, [effect] may still occur because [evidence, purpose, permission or ordering] is unavailable or unenforced at [boundary/time].'' An LLM reviewer may draft such a record, but reported handoffs must refer to real evidence and policies still require engineering judgment.

\begin{center}
\begin{minipage}{.96\linewidth}
\setlength{\fboxsep}{9pt}\fcolorbox{edge}{pale}{\begin{minipage}{.94\linewidth}\small
\textbf{TRACE: a premise-to-effect review card (a chain, not a score)}\par\smallskip
\textbf{PICK} one human-facing effect subject to delay, changing context, handoff or significant consequence.\\
\textbf{RECORD} T: trigger; R: explicit operation versus purpose; A: evidence, source and age; C: authority and ordering; E: execution versus achievement. Record \emph{producer $\rightarrow$ artifact $\rightarrow$ decision owner $\rightarrow$ time $\rightarrow$ effect}.\\
\textbf{CHANGE} one premise, keeping the original request fixed. Delay suggests goal/timing; uncertain observations suggest attention/provenance; handoffs suggest provenance/authority.\\
\textbf{FOLLOW} the change toward the effect:\\
\quad Q1: Under a stated policy, should the effect change?\\
\quad Q2: Was the premise established (absent, or available but unused)?\\
\quad Q3: Does acquired evidence and required purpose reach the decision boundary in time?\\
\quad Q4: Can that boundary alter, defer or sequence the effect?\\
\quad Q5: Would the remedy cause the opposite error?\\
\textbf{FIND} the earliest demonstrated break or record \emph{unknown}; a Q1 policy ambiguity calls for clarification, not a defect label.\\
\textbf{DECIDE} clarify policy, instrument first, repair/validate a boundary, or report \emph{no gap}. State paired acceptance scenarios and an inverse-risk check.
\end{minipage}}\end{minipage}
\end{center}

\textbf{An external example: a delayed medication reminder.} In a published social-robot study \cite{merino2025bt}, a robot is told to navigate to a bathroom to remind a person about medication. A closed door causes navigation failure, so the robot replans and takes a detour. Consider a change \emph{during that delay}: the person has already taken the medication. For this review, policy $P_m$ requires the robot to recheck the continued relevance of a materially delayed reminder before delivery, for example by asking if no sufficiently recent evidence is available. A reported taken, not-taken or unknown status calls for the response specified by the applicable care plan, not automatic cancellation. The published behavior tree does not check relevance before speaking. The public source code, released after the paper, contains a question-asking action and an answer condition that \emph{could} have been invoked, but were not in the illustrated tree \cite{merino2025code}. Thus the earliest evidenced break is \emph{Q2-unused}: the means to seek the information exist, but are not used. Independently, a failure-triggered repair path cannot by itself catch a \emph{moot success}---successful execution of an effect whose purpose has expired.\footnote{The published system is described in \cite{merino2025bt}; the later repository snapshot is commit \texttt{477bfcc}, 8 October 2025 \cite{merino2025code}. The illustrated reminder is a generic speech action, while the available question and condition primitives could be composed into a separate relevance check. Neither establishes the deployed configuration or any adverse outcome.}

The chain can also find \emph{no gap}: our approved handler repertoire blocked unapproved model-suggested actions. Table~\ref{tab:trace_matrix} contrasts the cases. Under review policy $P_v$, the visitor notification is reconsidered if reliable evidence indicates that the host has met the guest; $P_m$ requires a pre-delivery relevance check for a delayed medication reminder. Neither policy describes the deployed systems.

\begin{table*}[!t]
\centering\footnotesize\renewcommand{\arraystretch}{1.22}
\caption{TRACE dispositions across historical incidents and a published behavior-tree example; not a defect-detection benchmark.}
\label{tab:trace_matrix}
\begin{tabularx}{\linewidth}{@{}p{.25\linewidth}p{.09\linewidth}p{.15\linewidth}p{.16\linewidth}p{.13\linewidth}>{\raggedright\arraybackslash}X@{}}\toprule
\textbf{Effect / changed premise}&\textbf{Q1}&\textbf{Q2}&\textbf{Q3}&\textbf{Q4}&\textbf{Disposition}\\\midrule
Visitor already met & Y/$P_v$ & ?: encounter evidence & N: no purpose record & N: no review & Gap under $P_v$ \\
``No worries'': turn precedes reassurance & Y & Y: observed delay & Y: two paths & N: order & Observed sequencing gap \\
LLM proposes unapproved handler & Y & Y & Y & Y: blocked & No gap for this control \\
Medication status changes in transit \cite{merino2025bt} & Y/$P_m$ & N: never asked & -- & N: failure-only repair & Q2-unused in public example; deployment unknown \\
\bottomrule
\end{tabularx}
\par\vspace{3pt}\parbox{\linewidth}{\footnotesize\raggedright Y/P: yes under the stated policy; Y: supported; N: absent on the reviewed path; ?: unknown; --: not inferred after an upstream break. A downstream N may indicate an independently documented limitation, not proven information propagation.}
\end{table*}

The minimum proposed repair for the medication example is a \emph{policy-scoped generation and validation contract}: a purpose-dependent delayed effect should have a fresh, meaningful relevance check or an approved exception. The generator could insert a question/condition branch; an existing behavior-tree tester could then verify that an appropriate guard is present. Robust checking needs an identifiable action type or purpose tag, not merely a generic speech-node name. A reported dose is evidence, not proof of ingestion; verification should cover taken, not taken and unknown reports, and guard against unnecessary questions or inappropriate cancellation (Q5).

A complete, fillable TRACE worksheet, including a worked visitor record, is supplied as supplementary material.

\section{What to Preserve}
A reassurance spoken after Alice had already turned away, and a correct visitor announcement delivered after its likely purpose had expired, revealed how timing, evidence and retained purpose shape social meaning. Our ROS~2 toolkit made many of those responsibilities inspectable through named integration boundaries. Modern models can replace substantial portions of the old pipeline, offer richer observables, and generate more flexible skills. Their breadth does not automatically determine which observation is gathered, which task remains necessary, or which effect is permitted to proceed.

A system may hold the means to check yet never use them, and \texttt{Success} need not mean the purpose was served. TRACE turns this into a concrete review: specify a policy, perturb one premise, find the first evidenced break, and choose the smallest enforceable repair.

Future robot models and architectures will change, just as ours changed while we built it. Three questions should endure: \emph{Which understanding reaches the moment of action? Does execution success still serve the task's purpose? Can an authorized boundary check the evidence behind a proposed effect?} Whether a robot runs a VLA, a reasoning model, ROS experts or a hybrid, these responsibilities remain. \pullquote{Preserve accountability boundaries, not computational pipelines.}

\printbibliography
\end{document}